\documentclass[letterpaper]{article}
\usepackage{aaai}
\usepackage{times}
\usepackage{helvet}
\usepackage{courier}
\usepackage{xcolor}
\usepackage{amsmath,amssymb,amsfonts}
\usepackage[ruled, lined, linesnumbered]{algorithm2e}
\usepackage{tikz}
\usetikzlibrary{er,positioning}
\usepackage{graphicx}
\usepackage{textcomp}
\usepackage{xcolor}
\usepackage{numprint}
\npthousandsep{,}
\usepackage{multirow}
\usepackage{hyperref}
\usepackage{bbm}
\usepackage[inline, shortlabels]{enumitem}
\usepackage{todonotes}
\usepackage{pgfplots}
\usepackage{natbib}
\usepackage{bbm}
\usepackage{caption}
\usepackage{subcaption}
\usepackage{booktabs}
\pgfplotsset{width=0.35\textwidth}
\pgfplotsset{compat=1.11,
        /pgfplots/ybar legend/.style={
        /pgfplots/legend image code/.code={%
        \draw[##1,/tikz/.cd,bar width=2pt,yshift=-0.2em,bar shift=0pt]
                plot coordinates {(0cm,0.8em)};},
},
}
\begin{document}
%
\title{Uncertainty Aware Wildfire Management}
\author{Tina Diao\\
{Stanford University}\\
tdiao@stanford.edu
\And
Samriddhi Singla\\
{University of California, Riverside}\\
ssing068@ucr.edu
\And
Ayan Mukhopadhyay\\
{Stanford University}\\
ayanmukh@stanford.edu
\AND
Ahmed Eldawy\\
{University of California, Riverside}\\
eldawy@ucr.edu
\And
Ross Shachter\\
{Stanford University}\\
shachter@stanford.edu
\And
Mykel Kochenderfer\\
{Stanford University}\\
mykel@stanford.edu
}
\maketitle
\begin{abstract}
\begin{quote}
Recent wildfires in the United States have resulted in loss of life and billions of dollars, destroying countless structures and forests. Fighting wildfires is extremely complex. It is difficult to observe the true state of fires due to smoke and risk associated with ground surveillance. There are limited resources to be deployed over a massive area and the spread of the fire is challenging to predict. This paper proposes a decision-theoretic approach to combat wildfires. We model the resource allocation problem as a partially-observable Markov decision process. We also present a data-driven model that lets us simulate how fires spread as a function of relevant covariates. A major problem in using data-driven models to combat wildfires is the lack of comprehensive data sources that relate fires with relevant covariates. We present an algorithmic approach based on large-scale raster and vector analysis that can be used to create such a dataset. Our data with over 2 million data points is the first open-source dataset that combines existing fire databases with covariates extracted from satellite imagery. Through experiments using real-world wildfire data, we demonstrate that our forecasting model can accurately model the spread of wildfires. Finally, we use simulations to demonstrate that our response strategy can significantly reduce response times compared to baseline methods.
\end{quote}
\end{abstract}

\section{Introduction}

In 2018, a large wildfire (named \textit{Camp Fire}) in California resulted in the loss of 88 lives, displaced countless more, and destroyed more than 18,500 structures. The estimated cost of the destruction was a staggering \$15 billion. As we write this manuscript at the start of the wildfire season in the state, more than a million acres have already burned in California this year alone due to more than 7,000 wildfires. Wildfires have destroyed many towns and structures across the state. At one point in August 2020, the entire northern half of the state had been instructed to prepare for evacuation~\citep{CAFire}. Crucially, the time of the year that is referred to as the ``wildfire season'' in the state has only just begun.

Fighting wildfires is difficult. Rapid urbanization and the effects of climate change make urban and suburban areas particularly susceptible to wildfires. Known to be notoriously unpredictable, sudden changes in wind directions or weather can change the way fires spread~\citep{timeArticle}. Firefighters need to allocate limited resources in dynamic and uncertain environments to intervene and stop the spread of fire. 

The problem of combating wildfires is an example of a dynamic resource allocation. Such an approach to combat natural calamities like wildfires, floods, and earthquakes are difficult for several reasons. The dynamics of the events are unknown and complicated to model in closed-form. There are multiple covariates that affect the spread of fire like the vegetation, fuel, altitude, and wind, and the exact relationship between each covariate and fire spread is uncertain. Resource allocation is extremely challenging because of the difficulty of making prediction about the spread of fire and expected damage. Designing principled approaches to deploy resources is important to mitigate the effects of wildfires (and emergencies in general)~\cite{mukhopadhyay2020review}.

Events like fires affect large areas, and it is difficult to correlate large-scale data from various sources to analyze and study fires. To the best of our knowledge, there exists no comprehensive data source that combines fire occurrence with geo-spatial features, fuel levels, and weather to allow the research community to develop approaches to combat wildfires. Finally, the problem of deploying resources to manage wildfires is full of uncertainties. As fires spread, it becomes increasingly difficult to observe the true state on the ground due to the presence of smoke. As a result, first responders only have imperfect information to allocate their resources. 

\textbf{Contributions}: We make the following contributions in this paper. \textbf{1)} We present a decision-theoretic approach to dynamically allocate resources in uncertain environments to intervene against wildfires. Specifically, we model the response problem as a partially observable Markov decision process (POMDP) and present an approach to find optimal actions for suppression for a given state of the problem. Our approach accommodates the constraint that the \textit{true} state of the fire is difficult to observe in practice. \textbf{2)} Instead of focusing on \textit{physics-based} models, we present a data-driven approach that can simulate the spread of wildfires. We extract relevant covariates such as fuel levels, vegetation type, height of canopy, and elevation from satellite imagery to drive the simulation. \textbf{3)} The calculation of accurate zonal statistics is a bottleneck for large scale geo-spatial data analysis ~\citep{singla2018distributed}. We demonstrate how extremely large-scale geo-spatial data pertaining to wildfires can be combined with other covariates through large-scale distributed raster and vector analysis. Crucially, we release the dataset as open-source for the research community. \textbf{4)} Through experiments using real-world wildfire data from California, we demonstrate that our forecasting model can accurately model the spread of wildfires and our response strategy results in significant improvement in suppression efforts compared to baseline methods that do not consider potential fire spread. 
\section{Prior Work}

The dynamics of fire spread are usually modeled using \textit{physics-based models}. Popular fire spread models include ~\textit{BehavePlus}~\citep{andrews1986behave} and ~\textit{Farsite}~\citep{finney1998farsite}. They are based on mathematically modeling surface fire spread as a function of heat flux and fuel availability~\citep{rothermel1972mathematical}. Such models are widely used by first responders and fire fighters to forecast the spread of fires. A relatively modern approach is to predict the rate of spread by integrating real-time information about weather from sensors~\citep{altintas2015towards}. Data-driven modeling has also been used to model fire spread. Supervised machine learning techniques have been applied to uncover strong associations of factors to wildfire sizes and frequency using different data sources. For example, ~\citet{joseph2019spatiotemporal} investigated weather conditions and geographic characteristics of extreme fire patterns in contiguous United States and ~\citet{ghorbanzadeh2019spatial} examined wildfire susceptibility using geographic data in northern Iran. 

Response to wildfires traditionally uses simulation-based approaches to select locations of intervention that maximize the expected utility of suppression efforts. ~\citet{petrovic2012dynamic}model wildfire dynamics and examine the trade-off between multiple competing suppression efforts
to compute an optimal strategy for fire responses. Stochastic simulation and multi-agent coordination has also been explored to combat wildifires~\citep{fried2006analysing,martin2002optimization}.
~\citet{griffith2017automated} explore how suppression efforts can be optimized by solving a mixed-integer linear program and by using Monte-carlo approaches to find optimal actions in a Markov-decision process (MDP). Our approach to optimize suppression improves upon prior work~\citep{griffith2017automated} to address the uncertainty in state information. We also integrate a data-driven generative model to simulate the spread of fire to aid decision-making under uncertainty.
\section{Problem Description}

We consider a spatial area divided into a set of spatial cells $G$. Let $g_i\in G$ denote the $i$th cell. We represent the neighbors of a cell $g_i$ by $N_i$, for some definition of neighborhood (for example, neighbors of a cell can be the set of its adjacent cells). Consider that the total time in consideration is divided into $T$ time-steps.
We assume access to historical data of fire incidents $D$, which is a vector of tuples $\{(t_1,\ell_1,u_1,w_1),(t_2,\ell_2,u_2,w_2),\ldots,(t_n,\ell_n,u_n,w_n)\}$, where each incident $d_i \in D$ is identified by its time of occurrence $t_i$, location $\ell_i$ (mapping to a cell in $G$), intensity of fire observed $u_i\in \mathbbm{R}^{+}$, and a vector of spatio-temporal features $w_i \in \mathbbm{R}^m$. 
The features $w$ capture potential determinants of fire such as weather and the type of vegetation in a cell. 

\begin{figure}[h]
  \centering
    \includegraphics[height=5cm]{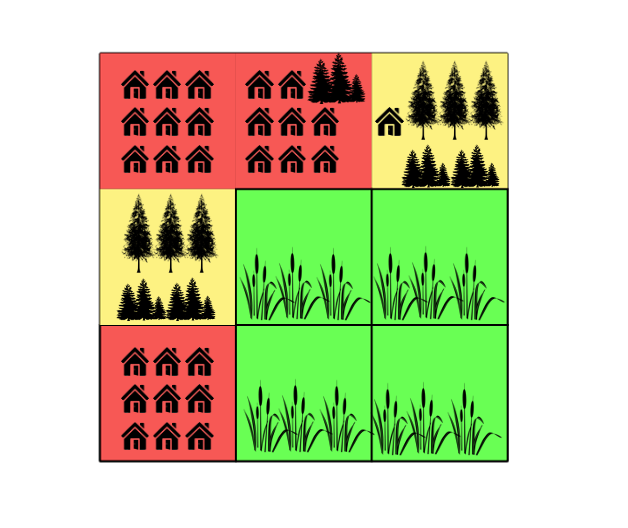}
    \caption{An example of a spatial area discretized in a grid with different costs and the colors correspond to varying costs to burn: red, yellow, and green in decreasing cost.}\label{fig:grid}
\end{figure}

Observing the true state of the world is almost always impossible when wildfires occur. Gathering real-time information by visiting the affected areas by land is naturally difficult. Therefore, information must be gathered through air surveillance obstructed by smoke. Consequently, the true dynamics of wildfires are only partially observable. We model the fire suppression problem more realistically as a partially observable Markov decision process (POMDP). A POMDP can be defined by the tuple $\{\mathcal{S}, \mathcal{A}, \mathcal{O}, Z, T, R, \gamma\}$~\citep{kochenderfer2015decision}. We define each component of the POMDP formulation below:

\begin{enumerate}
    \item States: $\mathcal{S}$ is a finite set of states. The state at time $t$ is denoted by $s_t = \{X_{t},F_{t}\}$, where $X_{t} = \{X^{1}_{t},X^{2}_{t},\dots,X^{k}_{t}\}$ and $F_{t} = \{F^{1}_{t},F^{2}_{t},\dots,F^{k}_{t}\}$ denote the status of the fire and fuel level in each of the cells in $G$, respectively. We consider the status of the fire $X$ as a binary variable such that
    \begin{equation*}
        x^{i}_{t} = \begin{cases}
        1 &\text{if $u^{i}_{t} \geq \epsilon$}\\
        0 &\text{otherwise}
        \end{cases}
    \end{equation*}
    where $u^{i}_{t} \in \mathbbm{R^+}$ denotes the measured intensity of the fire in cell $g_i \in G$ at time step $t$. We consider that the fuel level $F$ is a discrete variable, such that $F\in \{0,1,\dots,m\}$. The exogenous parameters $\epsilon$ and $m$ can be estimated from data or through domain knowledge. 
    \item Actions: $\mathcal{A}$ is a finite set of actions. The actions denote the different permutations of cell indices that fire suppression efforts can be applied to, up to a maximum number of cells specified as a resource constraint.
    \item State Transitions: $T$ defines conditional transition probabilities, with $T(s'\mid s,a)$ denoting the transition probability from state $s$ to $s'$ when action $a$ is taken. The transition model $T$ includes the following three components:

\begin{enumerate} 
	\item \textbf{Burning}: If cell $g_i \in G$ is burning at time step $t$, we assume that the fuel level decreases by one unit in the next time step. Therefore, $F^{i}_{t+1} = F^{i}_{t} - 1$ if $x^{i}_{t} \geq \epsilon$. 
	\item \textbf{Action effectiveness}: At any time step $t$, if an action, i.e. fire suppression effort(s) is applied to a cell $g_i \in G$ such that $x^i_t \geq \epsilon$, there is probability $q$ that the effort successfully puts out the fire. 
	\item \textbf{Fire dynamics}: We use a generative model to simulate the spread of fire. We represent spread dynamics by the probability distribution $f(X^{i}_{t+1} \mid x^{j}_{t}, w)$, where $g_i \in N_j$. Therefore, given that a specific cell $g_j \in G$ is on fire at time-step $t$, $f$ denotes the likelihood of its neighboring cells being on fire at the subsequent time-step $t+1$.
\end{enumerate}

We assume that in each time step, fire from a cell can spread only to its neighboring cells. This assumption aids computational tractability, but is without loss of generality because the decision-maker can discretize time fine enough such that the assumption is realistic. 

\item  Reward function: $R : \mathcal{S} \times \mathcal{A} \to \mathbbm{R}$ is the reward function, such that $R(s_t,a) = \sum_{g_i \in G} x^i_t U(g_i)$, where $U(g_i)$ denotes the (negative) utility for a cell $g_i \in G$ to be on fire. Naturally, $U$ varies across the cells. A cell with human occupants is presumably more \textit{valuable} than a cell composed of forested land. Cells can also carry unequal ecological utilities~\citep{bradshaw2012wildfire}. Without loss of generality, we consider three tiers of damage across cells, representing costs of residences (red), valuable ecological resource (yellow), and wildland (green) in decreasing values. Figure~\ref{fig:grid} shows an example grid with different types of cells. 

\item Observations and Observation Transitions: $\mathcal{O}$ represents the set of observations with $Z(o \mid s,a)$ denoting the probability of receiving observation $o$ at state $s$ when action $a$ is taken. We denote specific observations by $o^i_t$, which correspond to whether cell $g_i \in G$ is seen to be burning or not at time step $t$. Its transitions $Z(o \mid s, a)$ are deterministic, i.e. $o^i_{t,a} = u^i_t$ if an action has been applied to cell $g_i \in G$. This is because we assume to have ``eyes on location" when an action is applied to a cell. Otherwise, for cells where action for suppression is not applied (denoted by $\bar{a}$ in the expression below), a generative representation is used based on prior work~\citep{julian2019distributed} such that
\begin{align*} o^i_{t,\bar{a}} = 
	\begin{cases}
	1 & \text{if } Pr_t(X^i_{t}) > \gamma \text{ in state } s\\
	0 & \text{otherwise}
	\end{cases}
\end{align*}
where $\gamma$ is an exogenous parameter.

\item Discount factor: $\gamma \in [0,1]$ denotes the discount factor.
\end{enumerate}

In a POMDP, the decision-maker cannot directly observe the state. Instead, it only has access to beliefs that are generated probabilistically based on the actions taken. Information about states can be inferred from the history $(h)$ of observations and actions. It is common to maintain a distribution over states given the history; this distribution is known as the \textit{belief} $B$, such that $B(s \mid h)$ denotes the probability of being in state $s$ given history $h$. The goal for the decision-maker is to find a mapping from belief states to actions that maximizes the expected future discounted reward.



\section{Approach}
\subsection{Modeling Fire Spread}

In order to accurately simulate the spread of fire, we model the fire dynamics $f$ using a data-driven model. Recall that $f$ is a probability distribution over a cell being on fire in time step $t+1$, conditional on its neighbors being on fire at the previous time step $t$. 
The goal of modeling the function $f$ is to understand the effect of various covariates like wind, vegetation type, canopy height, altitude, etc. on fire spread. Typically, covariates in geo-spatial analysis are heterogeneous.

We use the random forests~\citep{liaw2002classification} which involves constructing a large number of decision trees at training time and then aggregating the outputs of the trees. The aggregation technique is typically using the mode of the outputs for classification and the mean of the outputs for regression. The central idea behind using random forests is to average many noisy but (approximately) unbiased models, thereby reducing the variance of the overall forecasting model.


\subsection{Resource Allocation}

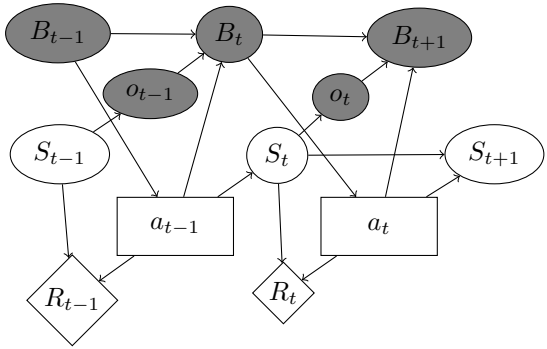
\begin{figure}
\centering
\begin{tikzpicture}[auto,node distance=0.4cm];
  \node[attribute, fill=gray] (node1){$B_{t-1}$};
  \node[attribute, fill=gray] (node4)[below right = of node1]{$o_{t-1}$};
  \node[attribute] (node2)[below left = of node4]{$S_{t-1}$};
  \node[entity] (node3) [below right = of node2] {$a_{t-1}$};
  \node[relationship] (rel1) [below left = of node3] {$R_{t-1}$};
  \node[attribute, fill=gray] (node5)[above right = of node4]{$B_t$};
  \node[attribute] (node6) [above right = of node3]{$S_t$};
  \node[entity] (node7) [below right = of node6]{$a_t$};
  \node[relationship] (rel2) [below left = of node7] {$R_{t}$};
  \node[attribute, fill=gray] (node8) [above right = of node6]{$o_{t}$};
  \node[attribute, fill=gray] (node9)[above right = of node8]{$B_{t+1}$};
  \node[attribute] (node10)[above right = of node7]{$S_{t+1}$};
  \draw [->] (node1) edge node {} (node3);
  \draw [->] (node3) edge node {} (node5);
  \draw [->] (node1) edge node {} (node5);
  \draw [->] (node2) edge node {} (node4);
  \draw [->] (node4) edge node {} (node5);
  \draw [->] (node2) edge node {} (rel1);
  \draw [->] (node3) edge node {} (rel1);
  \draw [->] (node3) edge node {} (node6);
  \draw [->] (node5) edge node {} (node7);
  \draw [->] (node7) edge node {} (node9);
  \draw [->] (node5) edge node {} (node9);
  \draw [->] (node6) edge node {} (node8);
  \draw [->] (node8) edge node {} (node9);
  \draw [->] (node6) edge node {} (rel2);
  \draw [->] (node7) edge node {} (rel2);
  \draw [->] (node7) edge node {} (node10);
  \draw [->] (node6) edge node {} (node10);
\end{tikzpicture}
\caption{POMDP (factored) representation of the dynamic decision problem. Shaded ovals reflect what is known to the decision maker.} \label{fig2}
\end{figure}

The general dynamic decision framework following the POMDP formulation is shown in Figure~\ref{fig2}. Recall that the sources of uncertainty are the current state of the fire ($S$) and the dynamics of the spread (driven by the set of covariates $w$). At each time step, the available set of actions ($\mathcal{A}$) is a combination of the fire location(s) to suppress. The decision-maker acts based on a belief distribution of the \textit{true} state of the world and receives an observation.  Utility (\textit{U}) reflects a measure of the expected damage of having a fire in a specific cell (e.g. residence versus wildland). 

Due to the large state space and action space (the action space is combinatorial), we use the sampling-based online Monte Carlo tree search (MCTS)~\citep{kochenderfer2015decision}. To address the computational complexity of POMDPs, approaches based on MCTS typically use a particle filter to represent beliefs in the search tree. Specifically, an approach that is of relevance to our problem is the partially observable Monte Carlo planning algorithm with observation widening (POMCPOW)~\citep{sunberg2017online}. POMCPOW differs from other online MCTS algorithms in that in the simulation step, given a state ($s$), history ($h$), and depth (for tree exploration), it weights the belief nodes and expands the belief updates gradually as more simulations are added. At each step, a single simulated new state is added to the particle collection, weighted to approximate the belief in every tree node, which is then used to sample the next new state. 



Although POMCPOW is shown to outperform other algorithms~\citep{sunberg2017online}, an issue with directly using POMCPOW to our problem is that the observation space is large and complex. This leads to a severe sparsity of particles, i.e., the probability of sampling a relevant observation is very small. To alleviate this, we modify the routine used to update belief in POMCPOW. Specifically, we replace the weighted particle filter with the standard \textit{particle filter without rejection}~\citep{kochenderfer2015decision}. As shown in Algorithm~\ref{particlefilter}, given a current belief ($b$), action ($a$), and observation $o$, $|b|$ samples are generated from the simulator, weighted, and subsequently resampled by its weights. Before the \textit{resampling} step, all weights of an observation $o$ may be 0 due to the large observation space.
If all the weights of an observation are 0, the probabilities of the sampled states are normalized to be proportional to the number of states already sampled. This reweighing step makes an approximation to importance resampling that seeks to estimate properties of a target distribution through sampling from a different distribution.

\begin{algorithm}

\SetKwInOut{Input}{Input}\SetKwInOut{Output}{Output}
\Input{Belief $b$, action $a$, observation $o$}
\Output{Updated belief $b'$}
\BlankLine
$b' \leftarrow \emptyset $\\
 \For{$i \leftarrow \{1,\dots,\mid b\mid\}$}{
  $s_i \leftarrow$ random state in $b$\\
  $s'_i \sim G(s_i, a)$\\
  $w_i \leftarrow O(o \mid s'_i, a)$
 }
 \If{$\sum_{i=1}^{\mid b\mid}w_i = 0$}{
  	$w_i$ = $\frac{1}{\text{len}(w)}$\tcp*{reweight states}
  }
 \For{$i \leftarrow \{1,\dots,\mid b\mid\}$}{
  Randomly select $k$ with probability proportional to $w_k$ \\
    Add $s'_k$ to $b'$
 }
 \Return normalized $b'$
 \caption{UpdateBelief $(b, a, o)$} \label{particlefilter}
\end{algorithm}

\begin{figure*}[h]
  \centering
    \includegraphics[height=4cm]{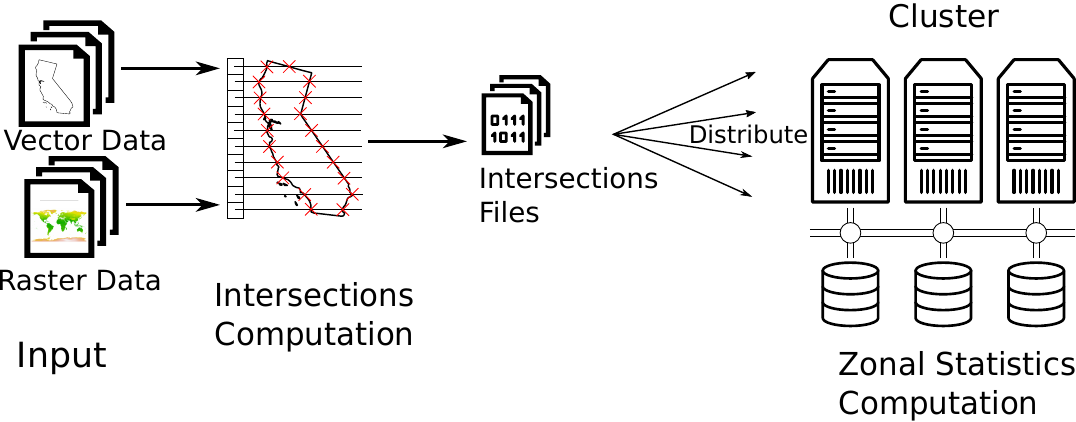}
    \caption{Architecture for calculating zonal statistics in large-scale raster and vector data}\label{fig:emi}
\end{figure*}

\subsection{Data Processing}
Data needed to model the spread of wildfires comes from varied sources. The temporal and spatial resolutions of such data sources are typically different. The data sources can also be of different forms (vector or raster). The vector model uses points and line segments to identify spatial locations while the raster model uses a set of cells for the same purpose. Combining large-scale vector and raster data is known to be a difficult problem~\citep{singla2018distributed}. 

We collected fire occurrence data in vector form from the Visible Infrared Imaging Radiometer Suite (VIIRS) thermal anomalies/active fire database~\citep{schroeder2014new}. The spatial resolution of VIIRS data is in the form of pixels representing 375 $\times$ 375 meter square cells~\citep{schroeder2014new}. The latitude and longitude values correspond to the center. Evidence of fire was read from the daily fire radiative power (FRP) levels in the VIIRS dataset. The data used to build the feature space was collected in raster form from the ``Landfire'' project~\citep{ryan2013landfire}. The foundation of the ``Landfire'' project is based on satellite imagery. The raster files had a spatial resolution of 30-meter square cells. This included features like canopy base desnity, canopy cover, and vegetation type. We list features used and the years from which the data was collected in Table~\ref{data}. 



\begin{table}[h]
\caption{LANDFIRE Raster Data} \label{data}
    \centering
    \begin{tabular}{ll}
    \toprule
    Name & Year(s) \\
    \midrule
    Canopy Base Density & 2012, 2014, 2016\\
    Canopy Base Height & 2012, 2014, 2016\\
    Canopy Cover & 2012, 2014, 2016\\
    Canopy Height & 2012, 2014, 2016\\
    Existing Vegetation Cover  & 2012, 2014, 2016\\ 
    Existing Vegetation Height & 2012, 2014, 2016\\
    Existing Vegetation Type & 2012, 2014, 2016\\
    Elevation & 2016\\
    Slope & 2016\\
    \bottomrule
    \end{tabular}
\end{table}

To reconcile the different spatial resolutions, we divide the state of California into a grid $G$ of 375 $\times$ 375 meter cells. The center of each fire pixel from the vector data can therefore overlap with exactly one cell in $G$. To compute the feature vector associated with each data point, we compute \textit{zonal statistics} for the vector data using the raster data. The method of zonal statistics refers to calculating summary statistics using a raster dataset within zones defined by another dataset (typically in vector form).


Traditional systems to compute zonal statistics require the data to be converted into the same format, either raster or vector. Automated tools can then be used to either vectorize the raster dataset or rasterize the vector dataset. The first approach converts each pixel in the raster to a point and then tests the point against each polygon in the vector data to find a match. This approach has a computational complexity of $O(n_p\log{n_p} \cdot c\cdot r)$, where $n_p$ is the number of polygons in the vector data, and $c$ and $r$ are the number of columns and rows in the raster data respectively.
The second approach rasterizes the vector data by converting each polygon to a raster (mask) layer with the same resolution as the input raster layer. It then combines the two raster layers to compute the desired aggregate function. Most systems that use this approach keep the mask layer in memory. If the input raster layer has a very high resolution, the size of the mask layer can become too large to be kept in memory.
This approach has a computational complexity of $O(n_p \cdot c \cdot r)$.

Neither of the two approaches mentioned above scale for the high resolution raster and vector data due to the requirement of converting between vector and raster formats. Specifically, in our problem, the vector data consists of over 3 million polygons and each of the raster data sources consists of over 1 billion entries. To deal with the large-scale geo-spatial data, we improve upon prior work ~\citep{singla2018distributed} to create a fully decentralized approach to compute large-scale zonal statistics. Our approach does not require data to be converted from one form to another (vector or raster). Instead, it computes an intermediate data structure, called an {\em intersections file} between the raster and vector data. The {\em intersections file} can be computed by using only the vector data and the metadata of the raster data (coordinate reference, resolution, etc.). Further, our approach can leverage parallel computation by using the intersection files. Our approach, with a computational complexity of $O(n_p\log{n_p} + c\cdot r)$, is scalable and efficient for large raster and vector datasets. Furthermore, as a by-product, our approach makes it easier to find the neighborhoods of polygons by performing a spatial self-join operation using the predicate intersects on all the cells in the vector data. We show the architecture of our approach in Figure \ref{fig:emi}.

\section{Experiments}

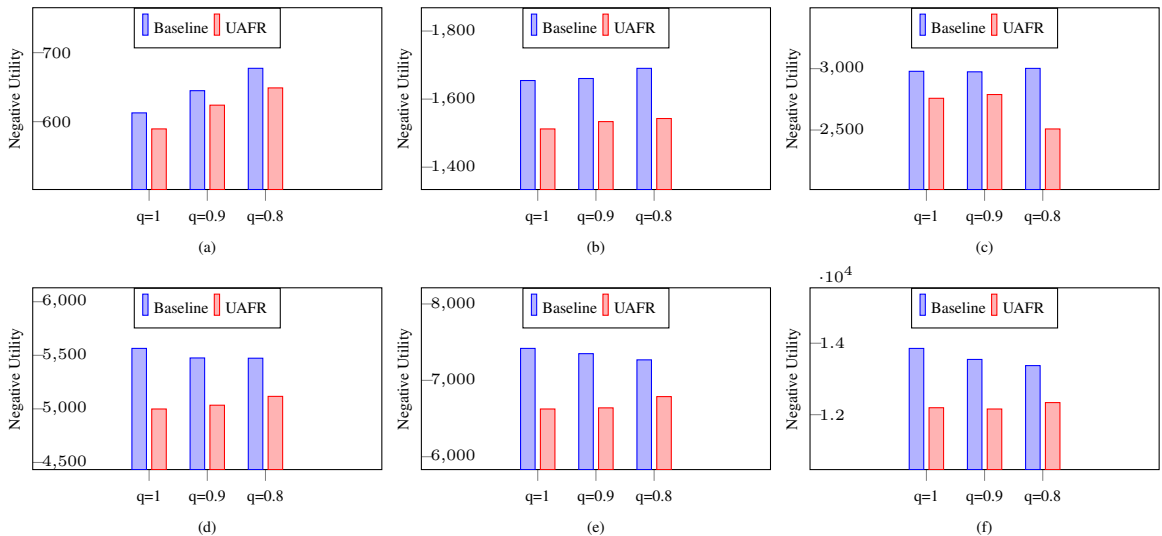
\begin{figure*}[h]
\centering
\begin{tikzpicture}
\label{fig:utlitycompare}
\begin{axis}[
    ybar,
    bar width=5.5pt,
    y tick label style={anchor=west},
    enlarge x limits=1,
    legend style={at={(0.5,1)},
      anchor=north,legend columns=-1,},
    ylabel={Negative Utility},
    title style={at={(0.5,-0.5)}},
    title=(a),
    symbolic x coords={q=1, q=0.9, q=0.8},
    xtick=data,
    x tick label style={ align=center},
    font=\tiny,
    xtick pos=left,
    ytick pos=left,
    enlarge y limits=1,
    height=4cm,
    ]
\addplot coordinates {(q=1,612.728) (q=0.9,644.91) (q=0.8,677.33)};
\addplot coordinates {(q=1,589.51) (q=0.9,623.926) (q=0.8,648.927)};
\legend{Baseline, UAFR}
\end{axis}
\end{tikzpicture}
\begin{tikzpicture}\label{fig:b}
\begin{axis}[
    ybar,
    bar width=5.5pt,
    y tick label style={anchor=west},
    enlarge x limits=1,
    legend style={at={(0.5,1)},
      anchor=north,legend columns=-1,},
    ylabel={Negative Utility},
    title style={at={(0.5,-0.5)}},
    title=(b),
    symbolic x coords={q=1, q=0.9, q=0.8},
    xtick=data,
    x tick label style={ align=center},
    font=\tiny,
    xtick pos=left,
    ytick pos=left,
    enlarge y limits=1,
    height=4cm,
    ]
\addplot coordinates {(q=1,1654.59) (q=0.9,1660.55) (q=0.8,1690.37)};
\addplot coordinates {(q=1,1512.33) (q=0.9,1533.87) (q=0.8,1542.95)};
\legend{Baseline, UAFR}
\end{axis}
\end{tikzpicture}
\begin{tikzpicture}\label{fig:c}
\begin{axis}[
    ybar,
    bar width=5.5pt,
    y tick label style={anchor=west},
    enlarge x limits=1,
    legend style={at={(0.5,1)},
      anchor=north,legend columns=-1,},
    ylabel={Negative Utility},
    title style={at={(0.5,-0.5)}},
    title=(c),
    symbolic x coords={q=1, q=0.9, q=0.8},
    xtick=data,
    x tick label style={ align=center},
    font=\tiny,
    xtick pos=left,
    ytick pos=left,
    enlarge y limits=1,
    height=4cm,
    ]
\addplot coordinates {(q=1,2978.44) (q=0.9,2974.16) (q=0.8,3002.26)};
\addplot coordinates {(q=1,2758.49) (q=0.9,2789.09) (q=0.8,2508.66)};
\legend{Baseline, UAFR}
\end{axis}
\end{tikzpicture}
\begin{tikzpicture}\label{fig:d}
\begin{axis}[
    ybar,
    bar width=5.5pt,
    y tick label style={anchor=west},
    enlarge x limits=1,
    legend style={at={(0.5,1)},
      anchor=north,legend columns=-1,},
    ylabel={Negative Utility},
    title style={at={(0.5,-0.5)}},
    title=(d),
    symbolic x coords={q=1, q=0.9, q=0.8},
    xtick=data,
    x tick label style={ align=center},
    font=\tiny,
    xtick pos=left,
    ytick pos=left,
    enlarge y limits=1,
    height=4cm,
    ]
\addplot coordinates {(q=1,5564.54) (q=0.9,5475.3) (q=0.8,5472.9)};
\addplot coordinates {(q=1,4998.94) (q=0.9,5034.31) (q=0.8,5116.78)};
\legend{Baseline, UAFR}
\end{axis}
\end{tikzpicture}
\begin{tikzpicture}\label{fig:e}
\begin{axis}[
    ybar,
    bar width=5.5pt,
    y tick label style={anchor=west},
    enlarge x limits=1,
    legend style={at={(0.5,1)},
      anchor=north,legend columns=-1,},
    ylabel={Negative Utility},
    title style={at={(0.5,-0.5)}},
    title=(e),
    symbolic x coords={q=1, q=0.9, q=0.8},
    xtick=data,
    x tick label style={ align=center},
    font=\tiny,
    xtick pos=left,
    ytick pos=left,
    enlarge y limits=1,
    height=4cm,
    ]
\addplot coordinates {(q=1,7418.26) (q=0.9,7349.33) (q=0.8,7268.2)};
\addplot coordinates {(q=1,6624.93) (q=0.9,6639.64) (q=0.8,6787.18)};
\legend{Baseline, UAFR}
\end{axis}
\end{tikzpicture}
\begin{tikzpicture}\label{fig:f}
\begin{axis}[
    ybar,
    bar width=5.5pt,
    y tick label style={anchor=west},
    enlarge x limits=1,
    legend style={at={(0.5,1)},
      anchor=north,legend columns=-1,},
    ylabel={Negative Utility},
    title style={at={(0.5,-0.5)}},
    title=(f),
    symbolic x coords={q=1, q=0.9, q=0.8},
    xtick=data,
    x tick label style={ align=center},
    font=\tiny,
    xtick pos=left,
    ytick pos=left,
    enlarge y limits=1,
    height=4cm,
    ]
\addplot coordinates {(q=1,13854.6) (q=0.9,13546.3) (q=0.8,13373.7)};
\addplot coordinates {(q=1,12195.0) (q=0.9,12159.3) (q=0.8,12338.8)};
\legend{Baseline, UAFR}
\end{axis}
\end{tikzpicture}
\caption{Effectiveness measured by negative utility on test data (lower is better) on varying grid sizes: (a) 4 $\times$ 4 (b) 6 $\times$ 6 (c) 8 $\times$ 8 (d) 10 $\times$ 10 (e) 12 $\times$ 12 and (f) 16 $\times$ 16.}
\end{figure*}

We used fire data from California, USA spanning 2012--2018 for the prediction modeling. We divided the state into a set of 375 $\times$ 375 meter cells. Our goal is to capture how fire spreads given an initial occurrence of fire. As a result, we only considered cells and days that exhibited the possibility of fire spreading from an existing fire. Specifically, each cell in our data has a fire in its neighborhood. Our data has a total of 2,367,209 data points. Each row in our data represents a cell $g_i$ at a specific time $t$, the neighboring cells of $g_i$ denoted by $N_i$, a set of spatial-temporal features $w_{it}$, and the status of fire $u_i^t$. To calculate the features for a specific cell $g_i \in G$ at a specific time step, we calculated summary statistics (minimum, maximum, median, sum, mode, and count of the feature values) using all raster cells within $g_i$. To the best of our knowledge, our data is the first comprehensive open-source dataset that combines fire occurrence with relevant covariates extracted from satellite imagery. The data is available at \url{https://wildfire-modeling.github.io}. We used data from 2012 to 2017 as our training set and data from 2018 as our test set. We set the time step for our experiments to a day, based on the minimum time fidelity of the VIIRS dataset. All experiments were run on an Intel Xeon 2.2 GHz processor with 125 GB of memory.

\subsection{Fire Spread}

We label a forecast as a true positive prediction when both the predicted fire intensity and the recorded fire intensity are greater than the pre-specified threshold $\epsilon$. We observe that the random forest regression model is insensitive to the number of trees used (5, 50, 100, and 500). We also observed similar accuracy across training and test sets. We also tested several realizations of $\epsilon$ to test the robustness of our forecasting approach on different types of fire. Our results show that while it is relatively difficult to predict spread from extremely \textit{small} fires ($\epsilon=0.5$), our forecasting model achieves high accuracy $(>90\%)$ in predicting spreads from relatively \textit{larger} fires. 
We summarize the results in Table~\ref{pred_accuracy}. 


\begin{table}[h]
\caption{Accuracy with 5-tree Random Forest Regression} \label{pred_accuracy}
    \centering
    \begin{tabular}{ll}
        \toprule
        FRP Threshold ($\epsilon$) & Accuracy on Test Set\\
        \midrule
        0.5 & 75\%\\ 
        1 & 79\% \\
        5 & 91\%\\
        \bottomrule
    \end{tabular}
\end{table}

\subsection{Fire Response}

Our goal is to create a pipeline for modeling response to wildfires by utilizing the data-driven model of fire spread to optimize response decisions. However, in this paper, we present experimental results based on a simple spread model. Specifically, we simulate the spread of fire using fixed environmental conditions (for example, we use a fixed wind direction and rate of spread in each simulation). We calculate the probability of fire spread following a simplified version of the deterministic fire dynamics model in prior work~\citep{griffith2017automated}. 


We conducted experiments on different sized square grids ($4^2, 6^2, 8^2, 10^2, 12^2$ and $16^2$). We varied the starting states by random initialization of fire maps and cell types. We experimented with 6 different initial states. For each initial state, we created 256 spread scenarios by varying wind and rate of spread. For each initialization, a fixed percentage ($10\%$) of cells were considered to be on fire. The proportion of red, yellow, and green cells were set to be 20\%, 30\%, and 50\% respectively, consistent with the distribution of land use in California~\citep{landuseUSDA}. 
In all experiments, we set the fuel level in each cell to 5, i.e., each cell takes 5 days to completely burn out and deplete its fuel.
We use a baseline model consistent with fire suppression strategies used in practice. Specifically, our baseline fire suppression targets the cell with the maximum utility which shows evidence of fire (through observations). 
Our open-source code is built using the POMDPs.jl framework~\citep{egorov2017pomdps} and is available at \url{https://github.com/wildfire-modeling/response_model}.

Figure 4 shows the mean negative utility averaged across start states and spread scenarios. We also vary the action effectiveness $q$ since in practice, suppression efforts do not always completely puts out wildfires. We see that our approach (referred to as ``Uncertainty Aware Fire Response'' or UAFR) consistently outperforms the baseline method in all scenarios with more significant improvements in larger grids and increased action effectiveness. 
\section{Discussion}

Wildfires have caused massive damage in the last decade. It is particularly difficult for first responders to combat wildfires due to difficulties in surveillance and scarcity of resources. In this paper, we created a data-driven forecasting model by extracting relevant determinants of fire spread through satellite imagery. Then, we created an approach to wildfire suppression that explicitly takes state uncertainty into account. We also created, to the best of our knowledge, the first comprehensive dataset on wildfires that combines historical fire data with relevant covariates. Our dataset, with over a million data points, and our codebase is open-source for the research community to use.

There are some limitations to our current work. While our forecasting model shows high accuracy, we observe that the random forest regression model is insensitive to the number of decision tress in the ensemble. As a result, simpler methods like classification and regression trees (CART) \cite{breiman1984classification} might result in better generalization to unseen data. Second, our approach to suppression needs to be integrated with the data-driven fire spread model. Finally, while we simulated wind for our experiments, our open-source dataset does not contain information about wind. We are currently incorporating hourly wind data from National Oceanic and Atmospheric Administration (NOAA)\footnote{https://www.climate.gov/maps-data/dataset/wind-roses-charts-and-tabular-data} with our fire data to develop a more comprehensive dataset.

{\small
\bibliographystyle{aaai}
\bibliography{references}

\begin{thebibliography}{}

\bibitem[\protect\citeauthoryear{Altintas \bgroup et al\mbox.\egroup
  }{2015}]{altintas2015towards}
Altintas, I.; Block, J.; De~Callafon, R.; Crawl, D.; Cowart, C.; Gupta, A.;
  Nguyen, M.; Braun, H.-W.; Schulze, J.; Gollner, M.; et~al.
\newblock 2015.
\newblock Towards an integrated cyberinfrastructure for scalable data-driven
  monitoring, dynamic prediction and resilience of wildfires.
\newblock {\em Procedia Computer Science} 51:1633--1642.

\bibitem[\protect\citeauthoryear{Andrews}{1986}]{andrews1986behave}
Andrews, P.~L.
\newblock 1986.
\newblock {\em BEHAVE: fire behavior prediction and fuel modeling system: BURN
  subsystem, Part 1}, volume 194.
\newblock US Department of Agriculture, Forest Service, Intermountain Research
  Station.

\bibitem[\protect\citeauthoryear{Bradshaw and
  Lueck}{2012}]{bradshaw2012wildfire}
Bradshaw, K., and Lueck, D.
\newblock 2012.
\newblock {\em Wildfire policy: Law and Economics Perspectives}.
\newblock Routledge.

\bibitem[\protect\citeauthoryear{Breiman \bgroup et al\mbox.\egroup
  }{1984}]{breiman1984classification}
Breiman, L.; Friedman, J.; Stone, C.~J.; and Olshen, R.~A.
\newblock 1984.
\newblock {\em Classification and regression trees}.
\newblock CRC Press.

\bibitem[\protect\citeauthoryear{Egorov \bgroup et al\mbox.\egroup
  }{2017}]{egorov2017pomdps}
Egorov, M.; Sunberg, Z.~N.; Balaban, E.; Wheeler, T.~A.; Gupta, J.~K.; and
  Kochenderfer, M.~J.
\newblock 2017.
\newblock {POMDP}s.jl: A framework for sequential decision making under
  uncertainty.
\newblock {\em Journal of Machine Learning Research} 18(26):1--5.

\bibitem[\protect\citeauthoryear{Finney}{1998}]{finney1998farsite}
Finney, M.~A.
\newblock 1998.
\newblock {\em FARSITE, Fire Area Simulator--model development and evaluation}.
\newblock Number~4. US Department of Agriculture, Forest Service, Rocky
  Mountain Research Station.

\bibitem[\protect\citeauthoryear{Fried, Gilless, and
  Spero}{2006}]{fried2006analysing}
Fried, J.~S.; Gilless, J.~K.; and Spero, J.
\newblock 2006.
\newblock Analysing initial attack on wildland fires using stochastic
  simulation.
\newblock {\em International Journal of Wildland Fire} 15(1):137--146.

\bibitem[\protect\citeauthoryear{Garza}{2020}]{timeArticle}
Garza, A.
\newblock 2020.
\newblock {AI} is helping fight wildfires before they start.
\newblock \url{https://time.com/5497251/wildfires-artificial-intelligence/}.

\bibitem[\protect\citeauthoryear{Ghorbanzadeh \bgroup et al\mbox.\egroup
  }{2019}]{ghorbanzadeh2019spatial}
Ghorbanzadeh, O.; Valizadeh~Kamran, K.; Blaschke, T.; Aryal, J.; Naboureh, A.;
  Einali, J.; and Bian, J.
\newblock 2019.
\newblock Spatial prediction of wildfire susceptibility using field survey gps
  data and machine learning approaches.
\newblock {\em Fire} 2(3):43.

\bibitem[\protect\citeauthoryear{Griffith \bgroup et al\mbox.\egroup
  }{2017}]{griffith2017automated}
Griffith, J.~D.; Kochenderfer, M.~J.; Moss, R.~J.; Mi{\v{s}}i{\'c}, V.~V.;
  Gupta, V.; and Bertsimas, D.
\newblock 2017.
\newblock Automated dynamic resource allocation for wildfire suppression.
\newblock {\em Lincoln Laboratory Journal} 22(2).

\bibitem[\protect\citeauthoryear{Joseph \bgroup et al\mbox.\egroup
  }{2019}]{joseph2019spatiotemporal}
Joseph, M.~B.; Rossi, M.~W.; Mietkiewicz, N.~P.; Mahood, A.~L.; Cattau, M.~E.;
  St.~Denis, L.~A.; Nagy, R.~C.; Iglesias, V.; Abatzoglou, J.~T.; and Balch,
  J.~K.
\newblock 2019.
\newblock Spatiotemporal prediction of wildfire size extremes with bayesian
  finite sample maxima.
\newblock {\em Ecological Applications} 29(6).

\bibitem[\protect\citeauthoryear{Julian and
  Kochenderfer}{2019}]{julian2019distributed}
Julian, K.~D., and Kochenderfer, M.~J.
\newblock 2019.
\newblock Distributed wildfire surveillance with autonomous aircraft using deep
  reinforcement learning.
\newblock {\em Journal of Guidance, Control, and Dynamics} 42(8):1768--1778.

\bibitem[\protect\citeauthoryear{Kochenderfer}{2015}]{kochenderfer2015decision}
Kochenderfer, M.~J.
\newblock 2015.
\newblock {\em Decision Making Under Uncertainty: Theory and Application}.
\newblock MIT Press.

\bibitem[\protect\citeauthoryear{Liaw, Wiener, and
  others}{2002}]{liaw2002classification}
Liaw, A.; Wiener, M.; et~al.
\newblock 2002.
\newblock Classification and regression by randomforest.
\newblock {\em R news} 2(3):18--22.

\bibitem[\protect\citeauthoryear{Martin-fern{\'A}ndez, Mart{\'\i}nez-Falero,
  and P{\'e}rez-Gonz{\'a}lez}{2002}]{martin2002optimization}
Martin-fern{\'A}ndez, S.; Mart{\'\i}nez-Falero, E.; and P{\'e}rez-Gonz{\'a}lez,
  J.~M.
\newblock 2002.
\newblock Optimization of the resources management in fighting wildfires.
\newblock {\em Environmental Management} 30(3):352--364.

\bibitem[\protect\citeauthoryear{Mukhopadhyay \bgroup et al\mbox.\egroup
  }{2020}]{mukhopadhyay2020review}
Mukhopadhyay, A.; Pettet, G.; Vazirizade, S.; Lu, D.; Baroud, H.; Jaimes, A.;
  Vorobeychik, Y.; Kochenderfer, M.; and Dubey, A.
\newblock 2020.
\newblock A review of emergency incident prediction, resource allocation and
  dispatch models.

\bibitem[\protect\citeauthoryear{Petrovic, Alderson, and
  Carlson}{2012}]{petrovic2012dynamic}
Petrovic, N.; Alderson, D.~L.; and Carlson, J.~M.
\newblock 2012.
\newblock Dynamic resource allocation in disaster response: Tradeoffs in
  wildfire suppression.
\newblock {\em PloS one} 7(4):e33285.

\bibitem[\protect\citeauthoryear{Rothermel}{1972}]{rothermel1972mathematical}
Rothermel, R.~C.
\newblock 1972.
\newblock {\em A mathematical model for predicting fire spread in wildland
  fuels}, volume 115.
\newblock Intermountain Forest \& Range Experiment Station, Forest Service.

\bibitem[\protect\citeauthoryear{Ryan and Opperman}{2013}]{ryan2013landfire}
Ryan, K.~C., and Opperman, T.~S.
\newblock 2013.
\newblock Landfire--a national vegetation/fuels data base for use in fuels
  treatment, restoration, and suppression planning.
\newblock {\em Forest Ecology and Management} 294:208--216.

\bibitem[\protect\citeauthoryear{Schroeder \bgroup et al\mbox.\egroup
  }{2014}]{schroeder2014new}
Schroeder, W.; Oliva, P.; Giglio, L.; and Csiszar, I.~A.
\newblock 2014.
\newblock The new {VIIRS} 375 m active fire detection data product: Algorithm
  description and initial assessment.
\newblock {\em Remote Sensing of Environment} 143:85--96.

\bibitem[\protect\citeauthoryear{Singla and
  Eldawy}{2018}]{singla2018distributed}
Singla, S., and Eldawy, A.
\newblock 2018.
\newblock Distributed zonal statistics of big raster and vector data.
\newblock In {\em Proceedings of the 26th ACM SIGSPATIAL International
  Conference on Advances in Geographic Information Systems},  536--539.

\bibitem[\protect\citeauthoryear{{State of California}}{2020}]{CAFire}
{State of California}.
\newblock 2020.
\newblock 2020 incident archive.
\newblock \url{https://www.fire.ca.gov/incidents/2020/}.

\bibitem[\protect\citeauthoryear{Sunberg and
  Kochenderfer}{2017}]{sunberg2017online}
Sunberg, Z., and Kochenderfer, M.
\newblock 2017.
\newblock Online algorithms for pomdps with continuous state, action, and
  observation spaces.
\newblock {\em arXiv preprint arXiv:1709.06196}.

\bibitem[\protect\citeauthoryear{{United States Department of
  Agriculture}}{2016}]{landuseUSDA}
{United States Department of Agriculture}.
\newblock 2016.
\newblock {FIA} state stats.
\newblock
  \url{https://www.fs.fed.us/pnw/rma/fia-topics/state-stats/California/index.php}.

\end{thebibliography}
}

\end{document}